\documentclass{article}
\usepackage{amsmath}
\usepackage{algorithm}
\usepackage{algpseudocode}
\usepackage{spconf,amsmath,graphicx} 
\usepackage{float}
\usepackage{placeins}
\usepackage{caption}
\usepackage{makecell}
\usepackage{lipsum}
\usepackage[fontsize=9pt]{fontsize}

\usepackage{color}

\usepackage{fix-cm}
\usepackage{balance}
\usepackage{enumitem}
\usepackage{amssymb}
\usepackage{multirow}
\usepackage{amsfonts}
\usepackage{bbding}
\usepackage[numbers,sort&compress]{natbib}

\usepackage{url}

\usepackage{booktabs}
\usepackage{adjustbox}

\let\OLDthebibliography\thebibliography
\renewcommand\thebibliography[1]{
  \OLDthebibliography{#1}
  \setlength{\parskip}{4pt} 
  \setlength{\itemsep}{-0.5ex} 
}

\usepackage{hyperref}
\hypersetup{
    colorlinks=true,
    linkcolor=blue,
    filecolor=magenta,      
    urlcolor=cyan,
}

\title{O\lowercase{vercoming} U\lowercase{ncertain} I\lowercase{ncompleteness} \lowercase{for} R\lowercase{obust} M\lowercase{ultimodal} S\lowercase{equential} D\lowercase{iagnosis} P\lowercase{rediction} \\ \lowercase{via} C\lowercase{urriculum} D\lowercase{ata} E\lowercase{rasing} G\lowercase{uided} K\lowercase{nowledge} D\lowercase{istillation}}

\name{Heejoon Koo}
\address{
University College London \\
\ heejoon.koo.17@alumni.ucl.ac.uk
}

\begin{document}

\maketitle

\begin{abstract}

In this paper, we present NECHO v2, a novel framework designed to enhance the predictive accuracy of multimodal sequential patient diagnoses under uncertain missing visit sequences, a common challenge in real clinical settings. Firstly, we modify NECHO, designed in a diagnosis code-centric fashion, to handle uncertain modality representation dominance under the imperfect data. Secondly, we develop a systematic knowledge distillation by employing the modified NECHO as both teacher and student. It encompasses a modality-wise contrastive and hierarchical distillation, transformer representation random distillation, along with other distillations to align representations between teacher and student tightly and effectively. We also propose curriculum learning guided random data erasing within sequences during both training and distillation of the teacher to lightly simulate scenario with missing visit information, thereby fostering effective knowledge transfer. As a result, NECHO v2 verifies itself by showing robust superiority in multimodal sequential diagnosis prediction under both balanced and imbalanced incomplete settings on multimodal healthcare data.

\end{abstract}


\section{Introduction}
\label{sec:intro}

Predicting future patient diagnoses, a.k.a. sequential (next visit) diagnosis prediction (SDP), based upon clinical records is crucial for enhancing healthcare decision-making \cite{koo2024next, choi2016doctor, choi2017gram, yang2021leverage}. Recent advances in multimodal learning, which integrate diverse modalities such as clinical notes and demographics, have significantly improved SDP accuracy \cite{yang2021leverage, koo2024next}. Nevertheless, most previous studies assume the availability of all data, which is often impractical due to privacy, equipment failures, and other uncertain factors \cite{koo2024next}. Encountering such situations presents a formidable challenge to healthcare analytics.

Missing data, a common issue in reality, exacerbates model performance \cite{koo2023survey}. Basic approaches, such as imputation by mean or excluding incomplete data instances, often fail to preserve true data distribution and result in information loss \cite{emmanuel2021survey}. Advanced statistical techniques, including Multivariate Imputation by Chained Equations (MICE) \cite{van2011mice}, show better efficacy, but their application in complex multimodal scenarios still remains challenging. Therefore, deep learning approaches such as reconstruction \cite{yoon2018gain, lee2019collagan, yuan2021transformer}, which impute missing features, and knowledge distillation (KD) \cite{hinton2015distilling}, which transfers teacher's knowledge on complete data to a student learning with incomplete data \cite{wang2020multimodal, lin2023missmodal}, have gained popularity.

KD has proven effective in model compression \cite{sanh2019distilbert, jiao2020tinybert, kim2021comparing} and other applications, such as tackling missing data. Wang et al. \cite{wang2020multimodal} employs modality-specialised teachers that migrate knowledge to a multimodal student. MissModal \cite{lin2023missmodal} employs geometric multimodal contrastive loss \cite{poklukar2022geometric} and distribution alignment loss on a combination of modality representations in a self-distillation manner \cite{xu2019data}. However, there is a research gap in systematically leveraging KD to alleviate the representation discrepancy in teacher-student under missing data. Furthermore, no existing methodologies have taken into account the fixed dominance of specific modalities under complete data and the fluctuating modalities importance under incomplete data, leading to sub-optimal performance. 

Meanwhile, some studies examine the impact of data augmentation \cite{feng2021survey} on KD \cite{wang2022makes, li2022role}. However, research on applying data augmentation during KD with incomplete data remains under-explored.

To this end, we propose NECHO v2, not only overcoming the challenges in multimodal SDP with imperfect data \textit{for the first time} but also tackling the aforementioned limitations. First, we modify the original NECHO to manage uncertain modality dominance in the presence of missing data. Second, we establish a systematic KD pipeline, including modality-wise contrastive and hierarchical distillation, followed by transformer random representation distillation, MAG distillation, and dual-level logit distillation, to fully transfer the teacher's semantic knowledge acquired from the perfect data. Lastly, we introduce a random data erasing on each visit sequence in a curriculum fashion, simulating missing visit to reduce data distributional gap and facilitate representation propagation. By doing so, NECHO v2 shows consistent predominance under both balanced and imbalanced imperfect data scenarios on MIMIC-III data \cite{johnson2016mimic}.

\section{Methodologies}
\label{sec:methods}

\subsection{Problem Statement}
\label{sec:problem}

\noindent\textbf{Multimodal EHR Data.}
A clinical record is a time-ordered sequence of visits $V_1, . . . , V_T$, where $T$ represents the total number of visits for any given patient $P$. Each visit $V_t$ at $t$-th admission consists of three components: $D$, demographics; $N$, a clinical note; and $C$, a set of diagnosis codes. Specifically, a medical ontology $ \mathcal{G} $ is utilised to structure diagnosis codes into three hierarchical levels: unique codes, category codes, and disease-typing codes, from leaf to node. Input and target are unique codes and category codes.

\noindent\textbf{Missing Visit Sequences.} To simulate real-world missingness, we randomly discard aforementioned components in each visit sequence, creating an $m$-modal dataset with $ 2^m - 1 $ missing patterns. Missing probabilities are balanced or imbalanced across modalities and kept the same during both training and inference phases.

\begin{figure*}
\centering
\includegraphics[width=0.8\linewidth]{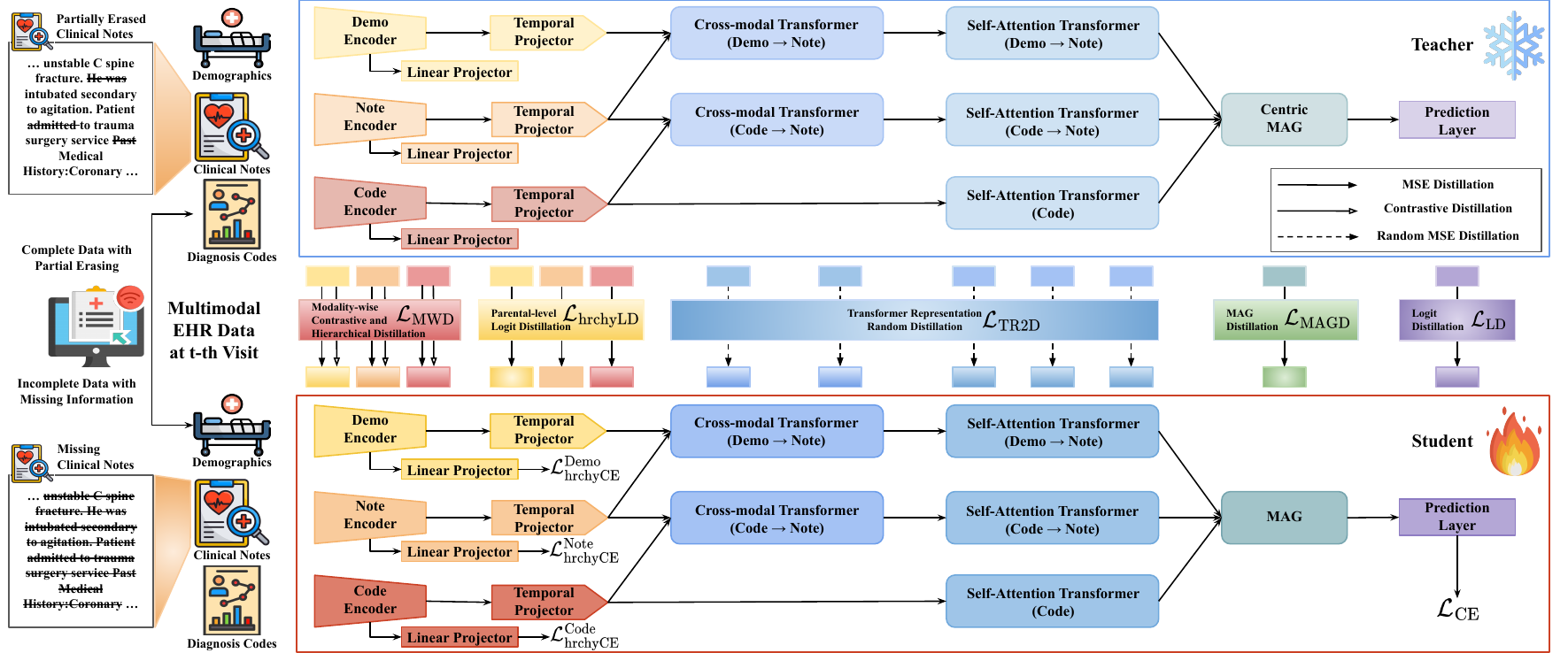}
\caption{A Visualisation of Our Proposed Framework, NECHO v2.
}
\label{fig:nechov2}
\end{figure*}

\noindent\textbf{Sequential Diagnosis Prediction.} Given a patient's incomplete multimodal clinical records for the past $ T $ visits, the objective is to predict diagnoses codes that will appear in the $(T + 1)$-th visit. 

\subsection{NECHO v2}
\label{sec:nechov2}

In this section, we present the KD-based NECHO v2 framework. For a comprehensive flow detailing the process from input to prediction, refer to the original NECHO \cite{koo2024next}.

\subsubsection{Modification of NECHO}
\label{sec:necho}

NECHO \cite{koo2024next} achieves state-of-the-art performance in SDP by integrating demographics, clinical notes, and diagnosis codes using a diagnosis code-centric framework with bi-modal contrastive loss and a centric multimodal adaptation gate (CMAG) for alignment and fusion. Each modality-specific encoder predicts at the parental level of target diagnosis codes (disease-typing codes) to enhance training.

However, it confronts two issues: 1) under-performance under incomplete data despite outstanding performance under complete data, and 2) adoption of a pre-trained BioWord2Vec \cite{zhang2019biowordvec}, limiting the adaptation to missing data. To mitigate these concerns, we modify by: 1) replacing the demo $ \rightarrow $ code transformers with a demo $ \rightarrow $ note transformers to relieve bias from diagnosis codes, and 2) utilisation of clinical TinyBERT \cite{rohanian2023lightweight} as a note encoder to potentially facilitate adaptability to incomplete data.

\subsubsection{Systematic Knowledge Distillation Framework}
\label{sec:systematickd}

\noindent\textbf{Teacher-Student Network Configuration.} In our KD pipeline, we adopt the modified NECHO as both teacher and student. Architecturally, the teacher leverages CMAG \cite{koo2024next} to consider modality representation dominance when learning with the full data, whilst the student adopts MAG \cite{yang2021leverage} that adjusts significant representations flexibly, considering fluctuating dominant features under missing conditions. Additionally, we avoid using the original NECHO as the teacher to reduce architectural heterogeneity, thereby fostering the KD \cite{hao2024one}. For the distillation process, we adopt offline distillation \cite{hinton2015distilling} where the teacher is trained, then frozen during distillation. Additionally, the teacher is absent during the student's inference.

\noindent\textbf{Modality-wise Contrastive and Hierarchical Distillation.} We begin our KD process by distilling modality-wise representations from the teacher to the student, using contrastive learning \cite{chen2020simple, radford2021learning} and L2 distance measures (Mean Squared Error, MSE). First, contrastive learning identifies and amplifies both similarities and discrepancies between the representations \cite{chen2020simple, radford2021learning, poklukar2022geometric}. When utilised in KD, it encourages the student's representations to be similar to those of the teacher's for corresponding samples, whilst also distinguishing between different samples. MSE further tightens this alignment, reducing deviations and promoting consistency. 

Unlike previous methods \cite{wang2020multimodal, lin2023missmodal}, we explicitly distill modality-specific semantic distributions. We utilise a contrastive loss with symmetrical losses to promote stable and effective distillation in a modality-wise fashion. Let teacher and student representations with the same data as positive sample pairs ($ \hat{R}_t^{\text{teacher}, m, i} $, $ \hat{R}_t^{\text{student}, m, i} $), with $ m \in (D, N, C) $, respectively. Then, with weighting parameter $ \alpha $ and batch size $ K $, the modality-wise contrastive distillation $ \mathcal{L}_{\text{MWCD}} $ is as follows:
\begin{equation}
    \mathcal{L}_{\text{MWCD}}^{\text{teacher $\rightarrow$ student}, m} = - \log \frac{\exp(\langle \hat{R}_t^{\text{teacher}, m, i}, \hat{R}_t^{\text{student}, m, i} \rangle/ \tau)}{\sum_{k=1}^K \exp(\langle \hat{R}_t^{\text{teacher}, m, i}, \hat{R}_t^{\text{student}, m, k}\rangle / \tau)},
\end{equation}

\vspace{-10pt}

\begin{equation}
    \mathcal{L}_{\text{MWCD}}^{\text{student $\rightarrow$ teacher}, m} = - \log \frac{\exp(\langle \hat{R}_t^{\text{student}, m, i}, \hat{R}_t^{\text{teacher}, m, i} \rangle/ \tau)}{\sum_{k=1}^K \exp(\langle \hat{R}_t^{\text{student}, m, i}, \hat{R}_t^{\text{teacher}, m, k}\rangle / \tau)},
\end{equation}

\vspace{-10pt}

\begin{equation}
    \mathcal{L}_{\text{MWCD}} = \sum_{m \in \{D, N, C \}} \{ \alpha \mathcal{L}_{\text{MWCD}}^{\text{\text{teacher},  $\rightarrow$ student}, m} + (1 - \alpha) \mathcal{L}_{\text{MWCD}}^{\text{student $\rightarrow$ \text{teacher}, }, m} \}
\end{equation}

where $ \langle, \rangle $ denotes cosine similarity and the temperature $ \tau \in \mathbb{R}^+ $ is a parameter that controls the distribution concentration and the gradient of the Softmax function. Next, with $ \hat{R}_t^{\text{teacher}, m} $ and $ \hat{R}_t^{\text{student}, m} $, modality-specific teacher and student feature at $t$-th visit, modality-wise hierarchical distillation $ \mathcal{L}_{\text{MWHD}} $ via MSE $ \| \cdot \|_2 $ is:
\begin{equation}
    \mathcal{L}_{\text{MWHD}} = \sum_{m \in \{D, N, C \}} \| \hat{R}_t^{\text{teacher}, m} - \hat{R}_t^{\text{student}, m} \|_2.
\end{equation}

Accordingly, the modality-wise contrastive and hierarchical distillation $ \mathcal{L}_{\text{MWD}} $ is formulated as the sum of the above two loss terms:
\begin{equation}
    \mathcal{L}_{\text{MWD}} = \mathcal{L}_{\text{MWCD}} + \mathcal{L}_{\text{MWHD}}.
\end{equation}

\noindent\textbf{Transformer Representation Random Distillation.} Previous research have explored intermediate layer distillation (ILD) between transformer layers for compression \cite{jiao2020tinybert, haidar2021rail, rohanian2023lightweight}. Meanwhile, NECHO has cross-modal (CMT) and self-attention transformer encoders (SAT) to align and merge inter- and intra-modality representations. Considering its multiple transformers, layer-wise distillation is computationally expensive. Hence, we distill teacher's randomly selected final transformer features, reducing computational burden and avoiding overfitting.

Denote representations from two CMTs as $ C^{M, D \to N}_t $ and $ C^{M, C \to N}_t $, and those from three SATs are $ S^{M, D \to N}_t $, $ S^{M, C \to N}_t $, and $ S^{M, C}_t $, where $ M $ is either teacher or student. Then, for the randomly selected transformer representations, the proposed distillation ($ \mathcal{L}_{\text{TR2D}} $) using MSE for both CMT ($ \mathcal{L}_{\text{CMTD}} $) and SAT ($ \mathcal{L}_{\text{SATD}} $) are:
\begin{equation}
    \mathcal{L}_{\text{TR2D}} = \mathcal{L}_{\text{CMTD}} + \mathcal{L}_{\text{SATD}},
\end{equation}

\vspace{-10pt}

\begin{equation}
\begin{aligned}
   \text{where}\, \mathcal{L}_{\text{CMTD}} = \sum_{m \in \{D, C\}} \| C^{\text{teacher}, m \to N}_t - C^{\text{student}, m \to N}_t \|_2,
\end{aligned}
\end{equation}

\vspace{-10pt}

\begin{equation}
\begin{aligned}
    \mathcal{L}_{\text{SATD}} = \sum_{m \in \{D, C\}} \| S^{\text{teacher}, m \to N}_t - S^{\text{student}, m \to N}_t \|_2 +  \\ \| S^{\text{teacher}, C}_t - S^{\text{student}, C}_t \|_2.
\end{aligned}
\end{equation}

\noindent\textbf{MAG Distillation.} To ensure the student model further mimics the teacher, we introduce MAG (penultimate layer) distillation. Its importance is also highlighted due to its rich, informative features \cite{wang2021distilling}. Let MAG representations from teacher and student be $ \text{CMAG}_t $ and $ \text{MAG}_t $, respectively. The regarding loss is:
\begin{equation}
    \mathcal{L}_{\text{MAGD}} = \| \text{CMAG}_t - \text{MAG}_t \|_2.
\end{equation}

\noindent\textbf{Dual Logit Distillation.} NECHO predicts target codes, as well as parental-level codes (disease typing codes) at the modality-specific encoders. Accordingly, we transfer both teacher predictions to the corresponding student predictions. Previous work \cite{kim2021comparing} argues that MSE outperforms Kullback-Leibler (KL) divergence for logit distillation, without requiring hyper-parameter tuning. Hence, MSE is applied to both distillations.

The final prediction and modality-specific parental-level prediction are $ \hat{y}^M_{t+1} $ and $ \hat{o}^{M, m}_{t+1} $. Then, the dual logit distillation loss $ \mathcal{L}_{\text{DualLD}} $ is written as:
\begin{equation}
    \mathcal{L}_{\text{DualLD}} = \mathcal{L}_{\text{LD}} + \mathcal{L}_{\text{hrchyLD}},
\end{equation}

\vspace{-10pt}

\begin{equation}
\begin{aligned}
    \text{where}\, \mathcal{L}_{\text{LD}} = \| \hat{y}^{\text{teacher}}_{t+1} - \hat{y}^{\text{student}}_{t+1} \|_{2}, 
\end{aligned}
\end{equation}

\vspace{-10pt}

\begin{equation}
\begin{aligned}
    \mathcal{L}_{\text{hrchyLD}} = \sum_{m \in \{D, N, C\}} \| \hat{o}^{\text{teacher}, m}_{t+1} - \hat{o}^{\text{student}, m}_{t+1} \|_{2}
\end{aligned}
\end{equation}

where $ \mathcal{L}_{\text{LD}} $ and $ \mathcal{L}_{\text{hrchyLD}} $ are final logit distillation and modality-specific hierarchical logit distillation, respectively. 

\noindent\textbf{Model Optimisation.} The student model is also optimised using a pair of task loss $\mathcal{L}_{\text{DualCE}}$ (CE stands for Cross Entropy), which consists of two components: one for the target level $\mathcal{L}_{\text{CE}}$ and the other for the parental level $\mathcal{L}_{\text{hrchyCE}}$, in accordance with NECHO.

By integrating the task losses with the aforementioned distillation losses with each constant $\lambda$, the full optimisation objective is formulated as:
\begin{equation}
\begin{aligned}
    \mathcal{L}_{\text{TOTAL}} = \lambda_{\text{MWD}} \mathcal{L}_{\text{MWD}} + \lambda_{\text{TR2D}} \mathcal{L}_{\text{TR2D}} + \\ \lambda_{\text{MAGD}} \mathcal{L}_{\text{MAGD}} + \lambda_{\text{DualLD}} \mathcal{L}_{\text{DualLD}} + \lambda_{\text{DualCE}} \mathcal{L}_{\text{DualCE}}.
\end{aligned}
\end{equation}

\subsubsection{Curriculum Learning Guided Random Data Erasing}
\label{sec:dataaug}

Prior study shows that large discrepancies in data distribution between teacher and student can hinder KD \cite{wang2022makes}. Therefore, we propose curriculum learning \cite{bengio2009curriculum} guided random single-point data erasing \cite{zhong2020random} to both training and distillation of the teacher. It is a minimalistic approach to mimic missing sequences and alleviate the data distribution gap to improve KD. Note that, it is not applied to the student during the distillation.

Firstly, the teacher is trained using curriculum learning guided random data erasing, starting with easier samples and gradually progressing to more difficult ones. All modalities are assigned a missing probability of 0.0 or 0.1 with equal probability until specific epoch $ e_1 $, after which the probability of 0.2 is added. Thereafter, during the distillation, complete data representations from the teacher trained in the previous manner are migrated until epoch $ e_2 $, after which training continues with either no missing data or a 0.1 missing ratio to each modality. 

This strategy improves robustness of the teacher against missing data during training and reduces data distribution discrepancies during distillation, leading to an improved representation transmission.

\section{Experiments}
\label{sec:experiments}

\subsection{Experimental Setup}
\label{sec:setup}

\noindent\textbf{Dataset and Pre-processing.} We evaluate on MIMIC-III data \cite{johnson2016mimic}, following pre-processing steps from previous works \cite{choi2017gram, koo2024next} but with a more rigorous patient selection criteria by removing records of: 1) with a length of stay of non-positive, and 2) who died within 30 days post-discharge. We also leverage only discharge summaries for clinical notes. Detailed statistics upon pre-processing are in Table \ref{tab:dataset_statistics}. 

To handle missing data, we assign a value beyond the existing range in demographics and diagnosis codes. For instance, if the total number of codes is 3882, the missing value is assigned as 3883. We also replace missing tokens in clinical notes with UNK token \cite{yuan2021transformer}.

\begin{table}[H]
\centering
\small
\scriptsize
\begin{adjustbox}{max width=\textwidth}
\begin{tabular}{ccc}
\toprule
\midrule
\textbf{Criteria} & \textbf{MIMIC-III} & \textbf{Count} \\
\midrule
\midrule
\multirow{4}{*}{General} & \# of Patients & 5551 \\
 & \# of Unique Codes & 3882 \\
 & \# of Category Codes & 126 \\
 & \# of Typing Codes & 17 \\
\midrule
\multirow{5}{*}{Visit} 
 & \# of Visits & 14568 \\
 & Avg / Max \# Visit per Patient & 3.37 / 33 \\
 & Avg / Max \# Unique Codes per Visit & 13.29 / 39 \\
 & Avg / Max \# Category Codes per Visit & 11.46 / 34 \\
 & Avg / Max \# Typing Codes per Visit & 6.71 / 15 \\
\bottomrule
\end{tabular}
\end{adjustbox}
\caption{Statistics of MIMIC-III Data After Pre-processing.}
\label{tab:dataset_statistics}
\end{table}

\noindent\textbf{Training and Evaluation Details.} We mostly follow the implementation details from previous study \cite{koo2024next}. We set the hidden dimension to 128 and the dropout rate to 0.1. The transformer encoders have 4 heads and 3 layers. We set the temperature $ T $ to 0.1 and the alpha $ \alpha $ to 0.25 for the contrastive distillation. The coefficients for loss terms are set to 1, except for the $\mathcal{L}_{\text{TR2D}}$ and $\mathcal{L}_{\text{hrchyCE}}$ which are 0.1. 

Optimisation is performed via AdamW \cite{loshchilov2017decoupled}, with a constant learning rate of 2e-5 for the parameters of clinical TinyBERT and 1e-4 for all other parameters. We train with a batch size of 4 for up to 100 epochs, stopping early if no improvement in validation set for 5 consecutive epochs. For curriculum learning, $ e_1 $ and $ e_2 $ are set to 5 and 10, respectively. 

\begin{table*}

\centering
\small
\begin{adjustbox}{max width=\textwidth}
\begin{tabular}{cccccccccccccccccccc}
\toprule
\midrule
\multirow{2}{*}{\textbf{Criteria}} & \multirow{2}{*}{\textbf{Models}} & \multicolumn{2}{c}{\textbf{(0.2, 0.2, 0.2)}} & \multicolumn{2}{c}{\textbf{(0.5, 0.5, 0.5)}} & \multicolumn{2}{c}{\textbf{(0.8, 0.8, 0.8)}} & \multicolumn{2}{c}{\textbf{(0.2, 0.2, 0.5)}} & \multicolumn{2}{c}{\textbf{(0.2, 0.8, 0.2)}} & \multicolumn{2}{c}{\textbf{(0.5, 0.2, 0.8)}} & \multicolumn{2}{c}{\textbf{(0.5, 0.8, 0.8)}} & \multicolumn{2}{c}{\textbf{(0.8, 0.2, 0.2)}}  & \multicolumn{2}{c}{\textbf{(0.8, 0.2, 0.8)}} \\
\cmidrule(r){3-4} \cmidrule(r){5-6} \cmidrule(r){7-8} \cmidrule(r){9-10} \cmidrule(r){11-12} \cmidrule(r){13-14} \cmidrule(r){15-16} \cmidrule(r){17-18} \cmidrule(r){19-20}
& & \textbf{top-10} & \textbf{top-20} & \textbf{top-10} & \textbf{top-20} & \textbf{top-10} & \textbf{top-20} & \textbf{top-10} & \textbf{top-20} & \textbf{top-10} & \textbf{top-20} & \textbf{top-10} & \textbf{top-20} & \textbf{top-10} & \textbf{top-20}  & \textbf{top-10} & \textbf{top-20}   & \textbf{top-10} & \textbf{top-20} \\
\midrule
\midrule
\multirow{4}{*}{Joint} & MulT \cite{tsai2019multimodal} & 35.52 & 51.83 & \underline{33.77} & 50.31 & \underline{30.27} & 46.74 & 34.23 & 50.74 & 33.82 & 50.06 & 32.71 & 49.41& \underline{30.39} & 47.29 & 36.01 & 52.78 &  \underline{33.58} & \underline{50.78} \\
& NECHO (Original) \cite{koo2024next} & 35.99 & 52.99 & 33.17 & 49.34 & 28.96 & 45.77 & 35.02 & 51.40 & 33.81 & 50.34 & 32.69 & 50.29 &  29.36 & 45.61 & 36.96 & \underline{53.46} & 31.60 & 48.62 \\
& NECHO (Modified for Teacher) & \underline{36.26} & 52.72 & 31.37 & 47.81 & 28.86 & 45.86 & 34.37 & 50.93 & 34.11 & 50.40 & 31.63 & 49.19 & 30.08 &  47.01 & \underline{36.85} & 53.21 & 33.20 & 50.29 \\
& NECHO (Modified for Student) & 35.96 & 52.98 & 33.24 & \underline{50.73} & 28.64 & 46.06 & \underline{35.29} & \underline{51.98} & 33.28 & 49.88 & \underline{33.28} & \underline{50.68} &  29.05 &  46.44 &  35.65 & 52.24 & 31.83 & 49.25  \\
\midrule
\multirow{3}{*}{KD} & UnimodalKD \cite{wang2020multimodal} & 35.45 & \underline{53.19} & 32.86 & 50.25 & 29.28 & 46.06 & 34.18 & 51.92 & \underline{34.32} & \underline{51.53} & 33.00 & 50.61 & 29.82 & 46.71 & 35.41  & 52.94  & 33.08 & 50.15  \\
& MissModal \cite{lin2023missmodal} & 35.85 & 52.80 & 33.41 & 50.37 & 30.00 & \underline{46.84} & 34.73 & 51.93 & 33.68 & 51.25 & 33.17 & 50.24 & 29.68 & \underline{47.34}  &  35.80 & 52.43  & 32.11 & 50.62  \\
& NECHO v2 (Ours) & \textbf{37.02} & \textbf{54.26} & \textbf{34.69} & \textbf{51.13} & \textbf{30.57} & \textbf{47.34} & \textbf{35.30} & \textbf{52.49} &  \textbf{34.73} & \textbf{51.65} & \textbf{34.24} & \textbf{51.01} & \textbf{30.87} & \textbf{48.07} & \textbf{37.41} & \textbf{53.84} &  \textbf{34.71} &  \textbf{50.94}  \\

\bottomrule
\end{tabular}
\end{adjustbox}
\caption{Experimental Results on Multimodal SDP with Uncertain Missingness on MIMIC-III Data. Missing ratios for each modality are ordered as: demographics, clinical notes, and diagnosis codes. Best results are in boldface and the second-best results are underlined.}
\label{tab:main_results}

\end{table*}

NECHO v2 is evaluated against joint learning methods (MulT \cite{tsai2019multimodal} and three NECHO \cite{koo2024next} variations: original, teacher, and student) and KD methods (UnimodalKD \cite{wang2020multimodal} and MissModal \cite{lin2023missmodal}). KD methods use the same teacher (or its encoders) and student for fair comparison. Evaluation uses top-$k$ accuracy with $k$ values of 10 and 20, following \cite{choi2016doctor, koo2024next}. Experiments are implemented using PyTorch \cite{paszke2019pytorch} and conducted on a single NVIDIA RTX A6000.

\subsection{Experimental Results}
\label{sec:results}

\subsubsection{Main Results}
\label{sec:main}

As shown in Table \ref{tab:main_results}, NECHO v2 demonstrates remarkable performance across various missingness scenarios on MIMIC-III dataset. Specifically, it outperforms MulT by 0.92\%, the original NECHO by 1.52\%, its teacher by 3.32\%, its student by 1.45\%, and UnimodalKD by 1.83\% in top-10 accuracy at the balanced missingness of 0.5. Similar trends are observed in other settings. 

In contrast, NECHO performs well when diagnosis codes are mostly present (0.2) but predicts poorly in scenarios where codes are highly missing (0.8). UnimodalKD and Missmodal underperform in most incomplete scenarios, highlighting the need for systematic knowledge distillation that accounts for fluctuating modality dominance under imperfect data. 

This remarkable performance gain of NECHO v2 is attributed to: 1) modifying NECHO to manage varying modality significance under imperfect data, 2) implementing systematic KD, including modality-wise contrastive and hierarchical distillation, to comprehensively mimic teacher at various representation levels, and 3) simulating random missing visit information by curriculum random data erasing to minimise data distribution gaps. These enables the student to imitate the teacher in varied incompleteness settings, ensuring considerable performance gain effectively.

\subsubsection{Ablation Studies}
\label{sec:ablation}

\begin{table}[h]
\centering
\small
\scriptsize
\begin{adjustbox}{max width=\textwidth}
\begin{tabular}{cccccc}
\toprule
\midrule
\multirow{2}{*}{\textbf{Criteria}} & \multirow{2}{*}{\textbf{Components}} & \multicolumn{2}{c}{\textbf{(0.2, 0.2, 0.2)}} & \multicolumn{2}{c}{\textbf{(0.5, 0.2, 0.8)}} \\
\cmidrule(r){3-4} \cmidrule(r){5-6}
 & & \textbf{top-10} & \textbf{top-20} & \textbf{top-10} & \textbf{top-20} \\
\midrule
\midrule
\multirow{4}{*}{KD} & w/o $ \mathcal{L}_{\text{MWCD}} $ & 37.10 & 54.10 & \textbf{34.37} & 50.79 \\
 & w/o $ \mathcal{L}_{\text{TR2D}} $ & 36.9 & 53.95 & 33.15 & 50.58 \\
 & w/o $ \mathcal{L}_{\text{MAGD}} $ & 36.01 & 53.27 & 32.91 & 49.87 \\
 & w/o $ \mathcal{L}_{\text{hrchyLD}} $ & 35.58 & 52.85 & 34.25 & 50.96  \\
\midrule
\multirow{3}{*}{DA} 
 & Only During Distillation & 36.42 & 53.32 & 34.05 & 50.83 \\
 & Only During Teacher Training & \textbf{37.28} & 53.65 & 32.71 & 49.72 \\
 & Not For Both & 36.43 & 53.68 & 33.54 & 50.93 \\
\midrule
NECHO v2 & Full & 37.02 & \textbf{54.26} & 34.24 & \textbf{51.01}  \\
\bottomrule
\end{tabular}
\end{adjustbox}
\caption{Ablation Studies on MIMIC-III Data.}
\label{tab:ablation_studies}
\end{table}

\noindent To evaluate our proposed components, we conduct ablation studies on MIMIC-III data, as detailed in Table \ref{tab:ablation_studies}. We report two scenarios: a balanced missing ratio of 0.2, and an imbalanced ratios of (0.5, 0.8, 0.2), representing two extremes where diagnosis codes representations are either highly dominant or minimal.

We first assess the effectiveness of KD. Whilst NECHO v2 occasionally performs better without $ \mathcal{L}_{\text{MWCD}} $ and $ \mathcal{L}_{\text{MWHD}} $, their consistent use generally enhances performance. The absence of $ \mathcal{L}_{\text{TR2D}} $ and $ \mathcal{L}_{\text{MAGD}} $ during distillation significantly deteriorates the performance, highlighting the importance of intermediate representation propagation. Additionally, $ \mathcal{L}_{\text{hrchyLD}} $ is beneficial. These validate the importance of all components in our systematic KD pipeline to align the student’s semantic knowledge to that of the teacher. 

We also evaluate the efficacy of data erasing against three scenarios: only during distillation, only during teacher training, and not for both. Overall performance considerably improves, highlighting the significance of the proposed curriculum random data erasing under missing visit information. This enhances the teacher's robustness against missingness during training and minimises data distribution discrepancies during distillation, resulting in the student model that is highly resilient to uncertain data incompleteness.

\subsubsection{Comparative Studies}
\label{sec:comparitive}

\begin{table}[h]
\centering
\small
\scriptsize
\begin{adjustbox}{max width=\textwidth}
\begin{tabular}{cccccc}
\toprule
\midrule
\multirow{2}{*}{\textbf{Criteria}} & \multirow{2}{*}{\textbf{Components}} & \multicolumn{2}{c}{\textbf{(0.2, 0.2, 0.2)}} & \multicolumn{2}{c}{\textbf{(0.5, 0.2, 0.8)}} \\
\cmidrule(r){3-4} \cmidrule(r){5-6}
 & & \textbf{top-10} & \textbf{top-20} & \textbf{top-10} & \textbf{top-20} \\
\midrule
\midrule
\multirow{2}{*}{Pairing} & Original $\to$ Original & 36.79 & 53.11 & 32.68 & 49.53 \\
 & Original $\to$ Modified for Student  & 36.44 & 52.96 & \textbf{34.66} & \underline{50.99} \\
\midrule
\multirow{1}{*}{$ \mathcal{L}_{\text{TR2D}} $} & Not Random  & \underline{36.91} & \underline{53.56} & 32.83 & 50.40 \\
\midrule
NECHO v2 & Proposed  & \textbf{37.02} & \textbf{54.26} & \underline{34.24} & \textbf{51.01} \\
\bottomrule
\end{tabular}
\end{adjustbox}
\caption{Comparative Studies on MIMIC-III Data.}
\label{tab:comparative_studies}
\end{table}

\noindent Under the same settings as the ablation studies, we compare our NECHO v2 with different teacher-student combinations (original to original, original to modified for student) and transformer not random distillation. Our proposed methodologies achieve the best overall performance, underscoring the importance of: 1) carefully pairing teacher and student to address shifting representation dominance and minimise architectural heterogeneity, and 2) incorporating randomness into KD to prevent overfitting. We provide the corresponding result to Table \ref{tab:comparative_studies}. 

\section{Conclusion}
\label{sec:conclusion}

We tackle uncertain missing sequences for robust multimodal SDP with the proposed NECHO v2.  With modified NECHO that dynamically adjusts dominant representations under varying missingness, we design a curriculum data erasing guided systematic KD pipeline that enables the student to effectively imitate the teacher. Extensive experiments on MIMIC-III data show the effectiveness of our approach over the existing methodologies. To foster future research, we release code at: \href{https://www.github.com/heejkoo9/NECHOv2}{https://www.github.com/heejkoo9/NECHOv2}.

\vfill\pagebreak
\ninept
\section{References}
\bibliographystyle{IEEEbib}
\bibliography{refs}
\vfill\pagebreak

\end{document}